\documentclass{article}

\usepackage[preprint,nonatbib]{nips_2018}

\usepackage[utf8]{inputenc} 
\usepackage[T1]{fontenc}    
\usepackage{hyperref}       
\usepackage{url}            
\usepackage{booktabs}       
\usepackage{amsfonts}       
\usepackage{nicefrac}       
\usepackage{microtype}      
\usepackage{listings}
\usepackage[utf8]{inputenc}
\usepackage{graphicx}
\usepackage{wrapfig}
\usepackage{amsmath,amsfonts,amssymb,amsthm}
\usepackage{subcaption}

\usepackage{color}
\usepackage{soul}  

\begin{document}

\title{Online Sensor Hallucination via Knowledge Distillation for Multimodal Image Classification}
\author{
  Saurabh Kumar,
  Biplab Banerjee,
  Subhasis Chaudhuri
  \\[8pt]
Indian Institute of Technology Bombay \\
\texttt{\{saurabhkm, bbanerjee, sc\}@iitb.ac.in}\\
}

\maketitle
\begin{abstract}
We deal with the problem of information fusion driven satellite image/scene classification and propose a generic hallucination architecture considering that all the available sensor information are present during training while some of the image modalities may be absent while testing. It is well-known that different sensors are capable of capturing complementary information for a given geographical area and a classification module incorporating information from all the sources are expected to produce an improved performance as compared to considering only a subset of the modalities. However, the classical classifier systems inherently require all the features used to train the module to be present for the test instances as well, which may not always be possible for typical remote sensing applications (say, disaster management). As a remedy, we provide a robust solution in terms of a hallucination module that can approximate the missing modalities from the available ones during the decision-making stage. In order to ensure better knowledge transfer during modality hallucination, we explicitly incorporate concepts of knowledge distillation for the purpose of exploring the privileged (side) information in our framework and subsequently introduce an intuitive modular training approach. The proposed network is evaluated extensively on a large-scale corpus of PAN-MS image pairs (scene recognition) as well as on a benchmark hyperspectral image dataset (image classification) where we follow different experimental scenarios and  find that the proposed hallucination based module indeed is capable of capturing the multi-source information, albeit the explicit absence of some of the sensor information, and aid in improved scene characterization.
\end{abstract}

\section{Introduction}
The current generation has witnessed the abundance of satellite RS images acquired by sensors with varied resolution (both spatial and spectral). Additionally, it is nowadays possible to obtain images of a given area on the ground almost on a daily basis, thus assisting in (near) real-time monitoring of the Earth's surface. Notice that images acquired by sensors with different characteristics showcase complementary information regarding the underlying scene which, if fused properly, is expected to better analyze the ground objects. Historically, the notion of RS scene recognition is considered to be the prime ingredient of a number of RS applications which basically associates semantic meanings to the respective scenes. Precisely, such a semantic tagging invariably helps in applications related to military, urban planning, disaster mapping and many more.

The task of scene recognition (\cite{haralick1973textural}) is inherently different from the classical problem of image classification in RS which aims at predicting semantic land-cover classes at the pixel level (\cite{janssen1992knowledge}, \cite{blaschke2001s}). However, the focus on regions or sites is more important than analyzing the scenes at the pixel level in a number of applications and in particular for images with a very fine spatial resolution where very limited information is contained in the individual pixels. Nonetheless, both the tasks can be accomplished under the availability of different levels of supervision, namely, fully-supervised, semi-supervised, and unsupervised. The RS literature is traditionally rich in classification algorithms which, by definition, work on features extracted from the pixels/images. Mostly, such features or a combination of a number of base features are extracted explicitly from the image under consideration, thus giving rise to the notion of uni-modal feature extraction.  Majority of the work in current literature follows this approach to address the classification problem (\cite{camps2014advances}, \cite{li2014review}).

On the other hand, with the availability of comparatively low-cost sensors and due to the successful launch of several international satellite missions (Sentinel 1 and 2, for example), multi-modal data are rather easier to obtain nowadays. Considering the complementary information captured by diverse sources (multi-spectral and PAN, for example), the performance of a given classifier system can substantially be boosted if this rich information can intuitively be combined since such a multi-modal fusion strategy is expected to enhance the discriminative capabilities of the input representations. As a naive example, while the panchromatic (PAN) images show better spatial resolution, it is understood that the corresponding multi-spectral (MS) images depicting the same ground area are rich in spectral properties. Hence, a weighted combination of both the modalities is always encouraged than considering each of them separately for a typical scene understanding task. Likewise, multi-spectral or hyperspectral imaging involves many measurements along the visible spectrum for each pixel and helps in coarse to fine level object analysis while synthetic aperture radar (SAR) observations are nearly insensitive to cloud cover and provide a completely different aspect regarding the back-scattering of the surface materials. 

Nonetheless, it is always expected that the individual image features which are likely to be fused should possess enough generalization proficiency. While for typical RS image classification, the spectral bands are in general considered as the feature descriptors, the extraction of image level features for scene recognition is non-trivial. The feature extraction paradigms have evolved from the hand-crafted ad-hoc features to the recent data-driven deep feature descriptors. Amongst the ad-hoc feature descriptors, SIFT, HOG, based local features are encoded with techniques like bag-of-visual-words, Fisher vector, VLAD etc. to obtain the image level representations (\cite{lowe2004distinctive}, \cite{dalal2005histograms}). On the other hand, global scene level descriptors (GIST) can directly accumulate the responses for a number of filters into a unified feature representation. Besides, primitive spectral and texture statistics like Gray level co-occurrence matrix (GLCM) and Local binary patterns (LBP) are also popular for RS images (\cite{su2008textural}, \cite{ojala2002multiresolution}). However, the advent of deep learning techniques has largely sidelined the ad-hoc descriptors for most of the visual inference tasks (\cite{lecun1995convolutional}, \cite{krizhevsky2012imagenet}, \cite{chen2014deep}). 

In particular, the deep convolutional neural networks (CNN) based descriptors are capable of learning very generic to very specific abstract feature hierarchy from the images and thus possess improved discriminative capability. Deep CNN based techniques, which can deal with both the single as well as multi-modal data are prevalent in the literature (\cite{vaduva2012deep}, \cite{hu2015transferring}, \cite{zhang2016deep}, \cite{romero2016unsupervised}). CNN based feature learning from a single modality is straight-forward as it involves the learning of a number of consecutive convolutional kernels in an end to end fashion. However, the design is non-trivial while dealing with multiple modalities. Apart from learning abstract features from multiple modalities in terms of fully-connected networks, attempts are made to augment multiple modalities prior to feeding the features to the classification networks. A number of recent endeavors have dealt with the problem of multi-modal feature fusion for RS images as mentioned in (\cite{gomez2015multimodal}).

However, the existing multi-modal fusion based CNN architectures require the presence of all the modalities during training and testing which may turn out to be a hard constraint for a number of applications particularly involving time-critical disaster management systems. Due to the variations in the temporal resolution of different sensors, it may not be possible to obtain all the different images immediately after a disaster takes place. Notwithstanding this limitation, the consideration of different images remains instrumental for efficient and quick response arrangement. One likely scenario in this regard can be formulated when \textit{all the modalities are present during training while one or more modalities are missing during testing or on field implementation}.

Inspired by the aforementioned discussions, we propose a deep CNN based hallucination model for multi-modal image fusion where the model is trained with all the modalities where the image level abstract feature descriptors are learned in an end-to-end fashion. We basically work in the privileged information setting where we have additional data available during training procedure only but not while testing. However, we handle the absence of part of the input modalities during testing by separately modeling a hallucination network which is capable of approximating the missing modalities from the available ones on the fly.
Ideally, the considered problem definition shares some ideological similarities with the classical missing value problem in RS. However, here we are interested in a more generic learning based scenario driven by the paradigms of modality hallucination and knowledge distillation jointly. To the best of our knowledge, there is an existing endeavor for RS modality hallucination (\cite{kampffmeyer2018urban}), but none on distilled hallucination. However as opposed to (\cite{kampffmeyer2018urban}), we judiciously incorporate the paradigm of knowledge distillation in our framework (see figure \ref{fig:alphaAbalation} for effect of distillation), apart from following a more intuitive training strategy. We summarize the noteworthy contributions of the paper in the following:

\begin{itemize}
    \item We propose a multi-modal fusion based scene recognition framework for RS data. Although we consider the presence of all the modalities during model training, some of the modalities are considered to be absent during testing. In order to approximate the missing modalities from the available ones, we design an intuitive deep hallucination network.
    \item  By incorporating the generalized distillation framework in the hallucination module, we demonstrate how an improved hallucination performance can be achieved.
    \item We conduct extensive experiments on a large-scale dataset consisting of multi-spectral and PAN image pairs for the purpose of scene recognition and on a benchmark hyper-spectral dataset for pixel classification. We observe that the distilled hallucination network is capable of producing recognition accuracies comparable to a scenario when all the modalities are present during testing. In addition, we also show how the learning capacity of the hallucination network varies with changes in various model parameters.
\end{itemize}

\section{Related work}
Considering the focus of the paper, here we discuss briefly on i) multi-modal fusion based scene recognition in RS, and ii) knowledge distillation based approaches in machine learning, and iii) idea of hallucination networks.

The usefulness of employing data from multiple modalities was demonstrated by early works (\cite{bell1995remote}, \cite{pohl1998review}) well before deep neural networks became popular. This included thermal image fusion with LiDAR data for object recognition (\cite{chu1992image}) along with scene interpretation (\cite{clement1993interpretation}) and multi-source image classification (\cite{solberg1994multisource}) with optical and SAR data fusion to mention a few. In remote sensing, multi-modal fusion has been used for problems including urban mapping (\cite{gamba2013human}), oil slick detection (\cite{brekke2005oil}), forest studies (\cite{dalponte2008fusion}) and disaster management (\cite{dell2012remote}). There are also works on generating other modalities like Digital Surface Models (DSM) and Digital Elevation Models (DEM) (\cite{remondino2006image}) to name a few.

The notion of \textit{two-stream networks} is popular in fusing multi-modal information:  RGB and optical flow image for video analysis 
(\cite{simonyan2014two}), Supervision transfer across RGB and depth (\cite{gupta2016cross}), spectral-spatial data integration in RS (\cite{hao2018two}), RGB and depth images for scene understanding (\cite{garcia2018modality}), to name a few. By design, the two-stream network architecture is comprised of two separate neural networks and outputs from each of them are fused using an appropriate fusion strategy. As expected, each branch of the two-stream network, or a stream, typically deals with instances depicting only one modality. Recently, a number of endeavors can be observed in the area of hyperspectral image analysis for efficiently integrating both the spectral and spatial information (\cite{hao2018two}).

Besides, \cite{hu2017fusionet} propose a two-stream convolutional network for urban scene classification using PolSAR and hyperspectral data. \cite{liu2018remote} use two steam network architecture for pan-sharpening using panchromatic in one stream and multi-spectral data in the other and later fuse the outputs using a concatenation and fusion layer. There have been other fusion approaches such as max, sum and average but all these can be subsumed by a learning based fusion strategy. It is found that the learning based feature fusion technique offers improved performance than the fusion strategies following max or average pooling. Recently \cite{audebert2018beyond} propose a deep learning based multi-scale approach towards multi-modal semantic labeling with LIDAR and multi-spectral data.

There have been previous works to address the missing modality problem using neural networks. \cite{tran2017missing} propose a data/modality imputation approach and this offline imputed dataset is used for training a classifier. We on the other hand propose a to hallucinate missing modality on the fly in an online fashion. \cite{ngiam2011multimodal} use all the modalities for feature learning step but use one modality for supervised training step after which the testing with that modality missing. We in contrast use all available modalities for training and only one for testing. \cite{wang2015deep} propose a autoencoder based architectures for learning representations for data and in turn maximize the correlation of features across modalities. The method cannot be trained end to end and has a network for representation learning with SVM or classifier being modeled henceforth on the projections. Our method is end-to-end trainable and due to this nature can be quickly fine tuned to new data for improved hallucination performance as per need.

Before commencing the discussion on the existing knowledge distillation approaches, we briefly mention the notion of model training with side information for better inference.
Precisely, training with privileged information consists of learning a student model in presence of additional information provided by a teacher model (\cite{vapnik2009new}). The idea is that the additional inputs provided by the teacher are expected to help the student to learn a better model. In the privileged information setting the additional inputs from the teacher are available to the student network only at the training time. At the test time the student network is expected to perform better inference pertaining to the initial teacher guided training process. \cite{luo2018graph} recently proposed a graph-based distillation method to learn in presence of side information for action detection and action recognition tasks in videos.

 \textit{Knowledge distillation} refers to the process of intelligently transferring the knowledge of an already trained, larger/complex learning module or an ensemble of several models to a rather smaller counterpart which is expected to be easier to comprehend and deploy. Knowledge distillation for machine learning models was first introduced by \cite{bucilua2006model} while \cite{hinton2015distilling} presented a more general approach towards distillation within the scope of a  feed-forward neural network and for the purpose of hand-written digit recognition. \cite{bucilua2006model} considers the operations directly on the logits obtained from the softmax layer for learning the smaller model and minimize the squared difference between the logits from cumbersome and the smaller model. \cite{hinton2015distilling} scale the final logits by a temperature parameter to obtain soft thresholds and these are used to train the smaller model. They also show that matching the logits as done by \cite{bucilua2006model}, is a special case of the technique they propose.
 
However, from the perspective of the distillation framework mentioned in \cite{hinton2015distilling}, the knowledge gathered by a deep network is not interpreted merely as the network parameters (weights and biases) obtained during training but also includes the final logits generated by the network corresponding to the inputs. Precisely, \cite{hinton2015distilling} propose a much general methodology for distillation of knowledge from a larger teacher network to a drastically smaller student network by using a parameter termed as the \textit{temperature}. Essentially, the output logits of teacher network are normalized with respect to the temperature parameter and subsequently the softmax operation is carried out on them in order to obtain a softer threshold which leads to an improved and more general knowledge transfer to the student network. The outputs of neural networks are typically class probabilities which are obtained by using the softmax function which converts the output logits $z_i$, computed for each class into a probability, by comparing it to other logits.

Consider a scenario where we have data from multiple modalities available during training, but while testing only one or a subset of these are available. \textit{Hallucination networks} are trained to learn to mimic the modality that is absent by training with the hallucination loss. This paradigm can readily be applied to RS applications considering that the missing data problem is quite frequent in many scenarios. However, there exists only one previous work on hallucination networks by \cite{kampffmeyer2018urban} which particularly hallucinates the missing Digital Surface Model (DSM) data in a remote sensing dataset and demonstrates improvement over an RGB based standard classification approach. However, the model is validated on a rather constrained scenario with limited samples and there has been no effort on knowledge distillation.

As opposed to the previous endeavor, ours is one of the initial approaches introducing the idea of knowledge distillation for missing data prediction in RS.
Combining the concept of distillation and privileged information we demonstrate a two-stream network that can be trained with two modalities and can hallucinate the missing modality at test time. Note that the data from the missing modality during testing is what we consider as the so-called privileged information in the training stage. We transfer the knowledge learned by the stream that is trained on the missing modality at test time to another network which is referred to as hallucination. The approach presented in this paper is influenced by the notion of generalized distillation framework \cite{garcia2018modality} which is considered for the purpose of human action recognition combining RGB and depth sensors. Apart from introducing the idea to the RS community, we specifically update their distillation framework with additional loss measure, which is showcased to boost the classification performance substantially.

\section{Proposed Approach}
\subsection{Preliminaries and Problem definition}
Let us consider a scenario where we have training data from multiple modalities: $(X_1, X_2, ... X_M)$, where $M$ denotes the total number of sensory modalities. In the classical supervised learning setting, one would prefer to design a classifier system by training the model with a subset or all of these $M$ features and subsequently evaluate the model on similar data items so as to preserve the consistency of the feature space. As opposed to this scenario, we aim to leverage data from all the $M$ available modalities to train a robust classifier but subsequently ensure that consistent or even improved generalization performance is guaranteed while only a subset of the $M$ modalities is available during testing. Precisely, let $X_i$ denote the available data samples from the $i^{th}$ modality with $Y$ being the corresponding class label while $X_j$ represents the data from missing sensor modality during testing considering that we deal with two different data sources in a given scenario. In this circumstances, we are interested in learning three functional mappings: i) $f_{comb}: \{X_i,X_j \rightarrow Y \}$ for the two-stream model, ii) $f_{hall}: X_i \rightarrow X_j$ for hallucination network, and iii) $f_{class}: \{X_i, f_{hall}(X_i)\} \rightarrow Y$ for the final classification.

For notational convenience, let $\mathbb{X}=\{(X_i^k,X_j^k, Y^k)\}_{k=1}^N$ formally represent $N$ image-label pairs from $Y^k \in \{1,2,\ldots,C\}$ scene themes where each image is captured by two separate sensors showcasing two completely different perspectives: $X_i$ and $X_j$ may denote the multi-spectral and PAN images for a given area or a pair of non-overlapping subsets of the original set of bands in hyperspectral image. For the next section, since we experiment with two data modalities, we assume $X_1$ and $X_2$ to represent panchromatic and multi-spectral data respectively and the network stream working with them are termed as PAN-Net and MS-Net for brevity. Note that we describe the proposed model using this setup, however, needless to mention, $X_i$ and $X_j$ may represent any different information sources in general. In the following, we describe the training process followed under generalized distillation for the proposed pipeline.

\begin{figure}
\begin{centering}
\begin{subfigure}[b]{0.465\textwidth}
   \includegraphics[width=0.8\linewidth]{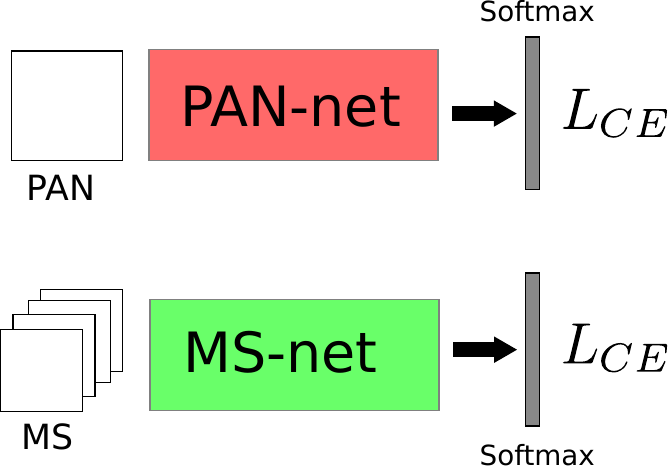}
   \caption{}
   \label{fig:step1} 
   \vspace{20pt}
\end{subfigure}
\end{centering}
\begin{centering}
\begin{subfigure}[b]{0.465\textwidth}
   \includegraphics[width=1\linewidth]{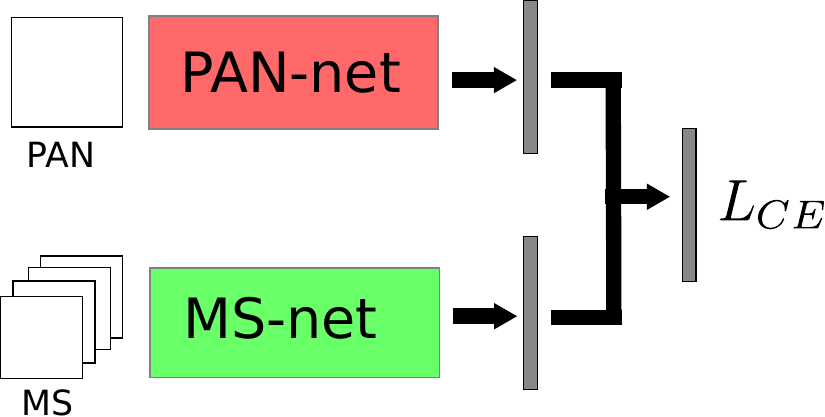}
   \caption{}
   \label{fig:step2}
   \vspace{20pt}
\end{subfigure}
\end{centering}
\begin{centering}
\begin{subfigure}[b]{0.465\textwidth}
   \includegraphics[width=1\linewidth]{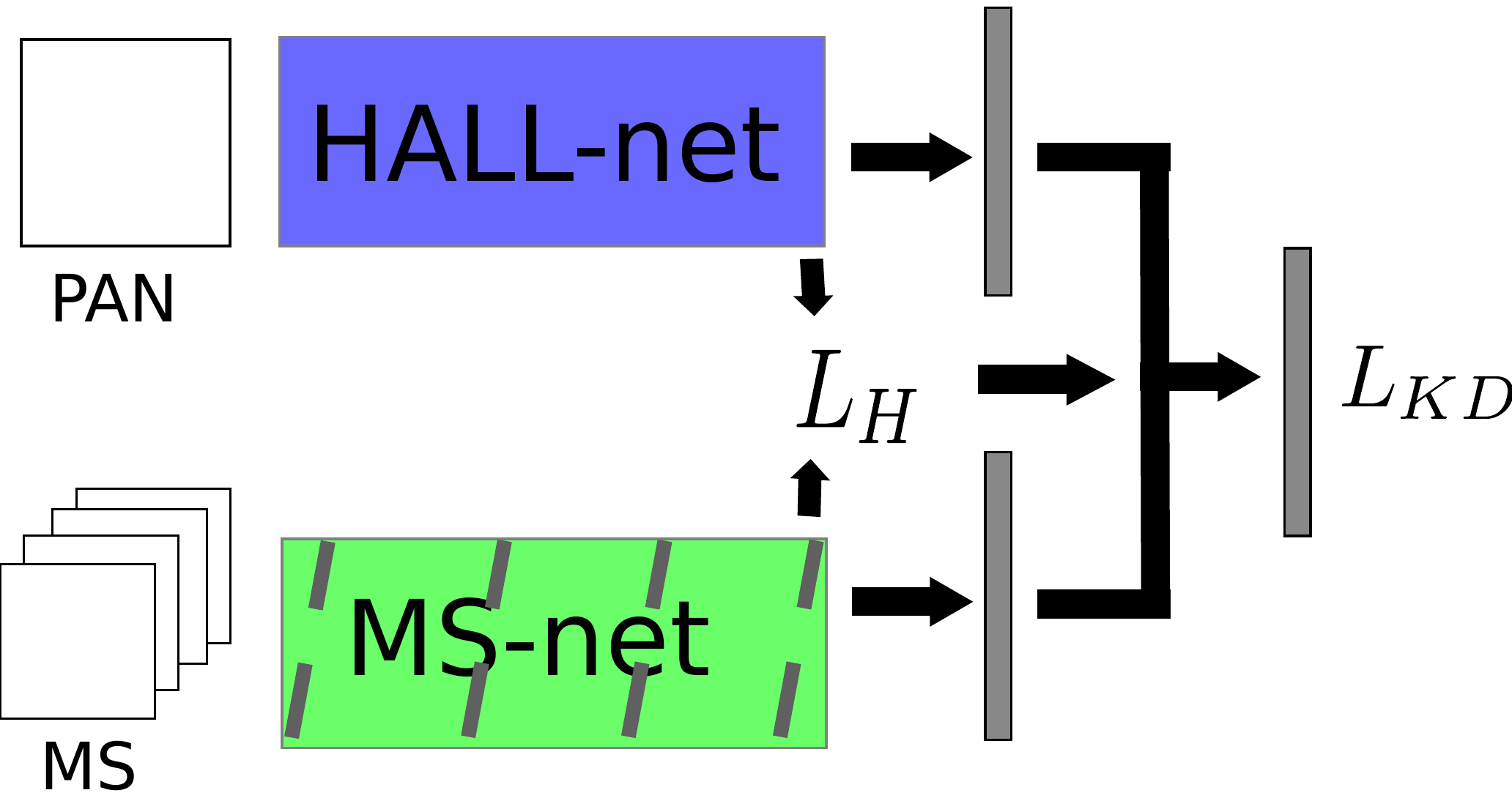}
   \caption{}
   \label{fig:step3} 
   \vspace{20pt}
\end{subfigure}
\end{centering}
\hspace{15pt}
\begin{centering}
\begin{subfigure}[b]{0.465\textwidth}
   \includegraphics[width=1\linewidth]{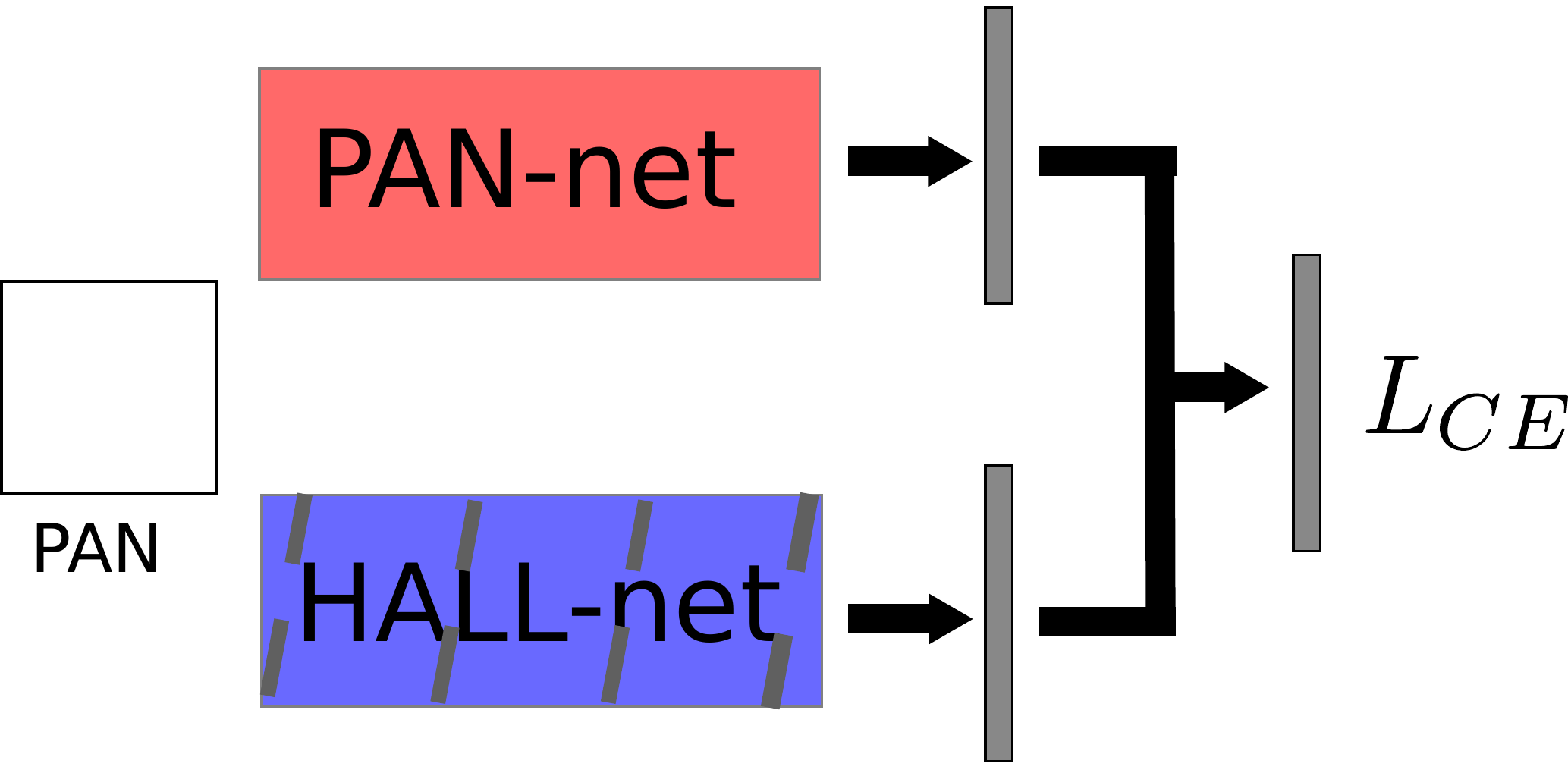}
   \caption{}
   \label{fig:step4}
   \vspace{20pt}
\end{subfigure}
\end{centering}

\caption{Training methodology as described in section 3.2, where multi-spectral data is unavailable during the testing phase. (a) Step-1: Both the networks are trained independently. (b) Step-2: Joint training of the two-stream network. (c) Step-3: Learning the hallucination network using the $L_{GD}$ and soft targets from frozen MS-Net, and finally (d) Step-4: Fine-tuning the hallucinated two-stream network with weights of the Hall-Net being frozen. The final step also represents how the network is deployed while testing. The striped networks represent that their weights are frozen in that step.}
\label{fig:trainingProcedure}
\end{figure}

\subsection{Training in detail}
We follow a four-step modular procedure for an intuitive training of our two-stream network. In particular, in the first two stages of training, we learn the \textit{teacher} network which is used in the following two steps in order to model the hallucination network. In the typical distillation setting, a \textit{teacher} network is the particular network whose outputs essentially guide the training of the second network which is termed as the \textit{student} network. Notice that with both $X_1$ and $X_2$ data streams being present, we initially learn separate network streams for each of the modalities followed by a joint learning combining both $X_1$ and $X_2$ in order to model $f_{comb}$. Subsequently, the stream for data modality that is known to be missing during testing is considered to be the teacher network and is deployed to learn the hallucination stream, $f_{hall}$ efficiently in conjunction with the generalized distillation loss. The trained $f_{hall}$, in turn, is used in the hallucinated two-stream network for classification which essentially has only the available modality as input and approximate the missing modality through $F_{hall}$ as if the testing is performed using both $X_1$ and $X_2$. The individual loss measures for each of the component networks are described in the following and Figure \ref{fig:trainingProcedure} depicts a flow for the training stages.

\subsubsection{Step-1: Learning the individual streams}
As aforementioned, we train each of the PAN-Net and MS-Net streams separately at the beginning. Essentially, these are designed as typical classification networks which perform classification solely based on the individual information source $X_1$ and $X_2$, respectively. Both of these networks are trained using cross entropy loss($L_{CE}$) for classification as given below:
\begin{equation}
    L_{CE}(Y^k, \hat{Y^k}) = - \sum_l Y_l^k \log\hat{Y_l^k}
\end{equation}
where, $Y^k$ and $\hat{Y^k}$ are the original and predicted labels in the typical one-hot vector fashion, respectively.

\subsubsection{Step-2: Learning the two-stream network}
\begin{figure}[!t]
\centering
\includegraphics[scale=0.8]{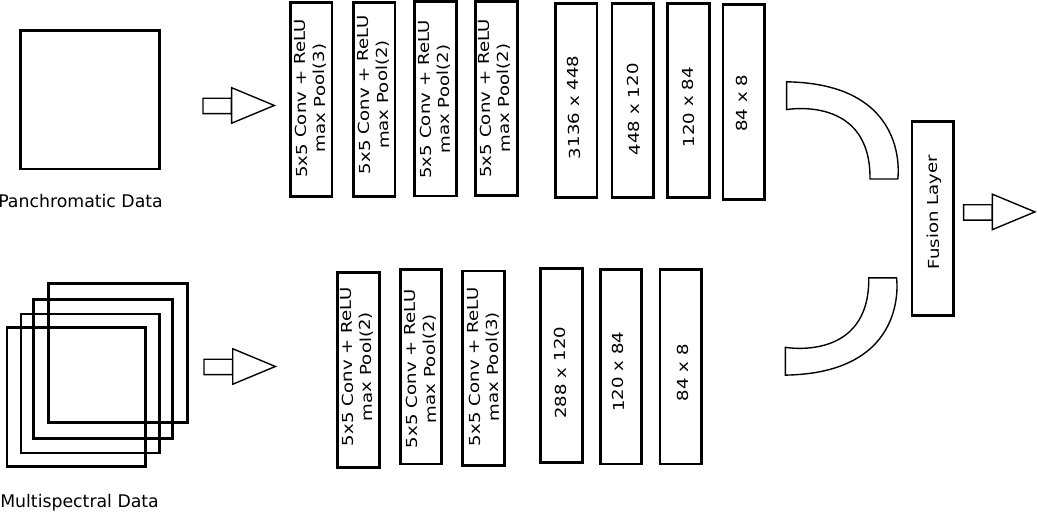}
\caption{Architecture of the two-stream network used for PAN-MS classification.}
\label{fig:fullArch}
\end{figure}

In the second phase of training, we build our two-stream network (Figure \ref{fig:fullArch}), $f_{comb}$ which is trained using data from both $X_1$ and $X_2$. The first stream uses the architecture of the PAN-Net and the second stream uses the architecture of MS-Net, respectively. We fuse the two streams at the end and the new representations are classified following a learned late fusion strategy. 

Specifically, the final class probabilities from both the streams are eventually concatenated and fed into a fusion layer. We employ a learned late fusion strategy, suppose $p_1, p_2 \in R^{N \times 1}$ are the probability score from the two streams respectively, we learn a set of fusion weights $w \in R^{N \times 2N}$, such that $ w *\!\ p = w_1 *\!\ p_1 + w_2 *\!\ p_2$, where $p \in R^{2N \times 1}$ is the concatenated probability vector. We want to learn a fusion layer such that,
\begin{equation}
    w = \arg\min_w \ell (p, Y; w) + \gamma \Vert w \Vert_2
\end{equation}
where $Y$ are the final labels and $\gamma$ is the regularization parameter. The symbol $ *\!\ $ represents the element-wise multiplication and $\Vert.\Vert$ is the standard $l_2$ norm.

For efficiently train the two-stream network, we first initialize the weights of the individual stream with the respective weights learned in Step-1 and the whole network is subsequently trained using cross-entropy loss. This strategy aids in better convergence in comparison to the random initialization.  This completes the learning of our teacher network for subsequent learning of hallucination network, which is the MS-Net for this description.

\subsubsection{Step-3: Learning the hallucination network}
In this step we train our hallucination stream, $f_{hall}$, using the generalized distillation process. Assume for instance that we wish to hallucinate the $X_2$ data stream while testing given data from $X_1$. In this case, we use the MS-Net architecture with data from $X_2$ as input which will act as a teacher network for the hallucination net which we alternatively term as the Hall-net. The Hall-Net or $f_{hall}$ follows the previous PAN-Net architecture and uses data from $X_1$ to generate corresponding the MS-Net activations. As shown in Figure \ref{fig:step3}, first we initialize the teacher network with the weights of MS-Net form Step-2 and freeze these weights. Next, we train the Hall-Net stream using the generalized distillation loss, to generate the multi-spectral counterparts for a given PAN instance.

In a standard knowledge distillation setting, we train the student model using a softer target distribution for the data samples via corresponding soft targets from the teacher network with the same high temperature in its softmax as the softmax of the student model. Let $z_i$ and $s_i$ be the logits and soft thresholds respectively, and $T$ is the temperature parameter. The outputs of the final layer in a neural network, before taking a softmax, are called the logits. The soft thresholds are obtained from the logits as follows,
\begin{equation}
s_i = \frac{exp(z_i/T)}{\sum_j exp(z_j/T)}.
\end{equation}
In the usual classification setting, the temperature parameter is typically set to $1$ to obtain the final probability distribution. But using a higher temperature value produce a softer probability distribution over classes which is employed during the knowledge distillation process.

Finally, this network is trained using the generalized distillation loss which comprises of the knowledge distillation loss term ($L_{KD}$) and hallucination loss term($L_H$) as given below. The hallucination loss is fundamentally a mean square error loss which minimizes the $l_2$ distance between the two soft thresholds, $s_{X_{2}}$ and $s_{Hall}$, of teacher and hallucination networks, respectively, which are obtained using a specific temperature.
\begin{equation}
    L_H = \Vert s_{X_2} - s_{Hall} \Vert^2
\end{equation}

The knowledge distillation loss, on the other hand, consists of two parts. Where first being the Kullback-Leibler divergence ($D_{KL}$) between the soft thresholds, $s_{X_{2}}$ and $s_{Hall}$. This is employed since we want to bring the two soft target distribution closer, allowing the student model to achieve better generalization. And the second part is a cross entropy loss between the hard labels, $z_{X_{2}}$ and $z_{Hall}$, from each of the two networks as we also want the hard target distribution to be as close as possible. The KL Divergence loss ($L_{KL}$) between the soft thresholds can be written as follows:
\begin{equation}
    L_{KL}(s_{X_{2}}, s_{Hall}) = D_{KL}(s_{X_{2}} \Vert s_{Hall})
\end{equation}

The complete knowledge distillation loss ($L_{KD}$) can now be formulated using the hyper-parameter $\lambda$ as follows:
\begin{equation}
    L_{KD} = \lambda T^2 L_{KL} + (1-\lambda) L_{CE}(z_{X_{2}}, z_{Hall}).
\end{equation}

Combining the hallucination and knowledge distillation loss we write the final generalized distillation loss ($L_{GD}$) as below:
\begin{equation}
    L_{GD} = \alpha L_{KD} + (1-\alpha) L_{H}
\end{equation}
where $\alpha$ is chosen to weigh in both these components suitably.

Note that the performance of the initial MS-Net stream is what we would ideally want the hallucination network to achieve eventually.

\subsubsection{Step-4: Learning the hallucinated(pseudo) two-stream network}
As the final phase of training and after having trained the $f_{hall}$ network to hallucinate the missing data stream $X_2$, we use it in the two-stream network, $f_{comb}$ from Step-2. We replace the missing modality stream of $f_{comb}$ with $f_{hall}$ to learn the final $f_{class}$ network. In particular, we initialize the first stream with weights from PAN-Net in Step-2 and the second stream with weights from Hall-Net learned in the previous step. Notice that the weights of the hallucination stream are frozen and subsequently we only fine-tune this hallucinated two-stream network with cross-entropy loss. This network is considered to be ready to perform classification with missing $X_2$ during decision making. Again, note that the performance of the initial two-stream network form Step-2 will be what we would ideally want to achieve with this hallucinated two-stream model.

\subsection{Testing}
The testing is straightforward given the hallucinated two-stream model from Step-4. We input $X_1$ in the PAN-net and $f_{hall}(X_1)$ in the HALL-net, respectively. The final classification scores are obtained by following the aforementioned late-fusion strategy mentioned in our original two-stream model (Step-2).

\section{Experimental Evaluations}
\subsection{Dataset description}
We use two datasets for analyzing the efficacy of the proposed hallucination model: the first one containing panchromatic images of aerial scenes along with the corresponding multi-spectral data. This dataset consists of a total of $80000$ PAN-MS pairs from eight different land-cover categories \cite{li2018learning}. The images are acquired by the GF-1 panchromatic and GF-1 multi-spectral sensors, respectively. The panchromatic images are of size  $128 \times 128$, with a spatial resolution of 2m and single spectral band. The respective multi-spectral data has a resolution of $64 \times 64$, with a resolution of 8m, and consists of a total of four spectral bands. The classes considered here are aquafarm, cloud, forest, high building, low building, farmland, river, and water. 

The second is the Pavia University hyperspectral datacube acquired by the ROSIS sensor\footnote{\url{www.ehu.eus/ccwintco/index.php/Hyperspectral_Remote_Sensing_Scenes}}. It is taken over the Pavia region in northern Italy and consists of $103$ spectral reflectance bands with $610 \times 340$ pixels each post discarding bands with unimportant information contents. The datacube bands have a geometric resolution of $1.3$ meters and the area contains pixels from nine land-cover classes namely, Asphalt, Meadows, Gravel, Trees, Painted metal sheets, Bare Soil, Bitumen, Self-Blocking Bricks and Shadows

\subsection{Model architectures}
Here we summarize the typical experimental setup considered for both the datasets. For experiments on the PAN-MS dataset, we alternatively consider the PAN and MS modality to be missing during testing. For the hyper-spectral data, we consider a subset of bands to be missing and aim at hallucinating them given the remaining bands.

For a given set of experiments conducted on the PAN-MS data for scene classification, we train one of the streams of our initial two-stream network with the panchromatic band and the second with multi-spectral data and provide only the panchromatic data for testing. Needless to mention, we focus to hallucinate the multi-spectral data for each test sample in this case. Notice that both of the network streams considered here are based on convolutional neural networks as our inputs are image objects. By design principle, the PAN-Net consists of four convolutional layers followed by four fully connected layers. Note that the structure is fixed heuristically.
On the other hand, the MS-net contains three convolutional layers and three subsequent fully-connected layers. In a separate set of experiments, we hallucinate the PAN images given the MS data-cubes.

Our network is inspired from the alexnet architecture \cite{krizhevsky2012imagenet} and since our dataset is relatively small we use a rather minimalistic version of the same. We tested the method with 3,4, and 5 convolution layers and find that the performances of the models do not drastically vary. Hence, we report the performance of the model with 3 hidden convolution layers. The late fusion scheme is standard in multi-stream deep learning models. One of our goals here was to keep the model simple and yet delivering good performance.

For the experiments concerning the hyper-spectral data, we use first $52$ bands (according to indices) of Pavia University datacube as our first sensor modality ($X_1$) and the goal is to hallucinate the subsequent $53$ bands during testing. Likewise in another set of experiments, we reverse the interpretation of both the streams and hallucinate the first $52$ bands while testing with only the last $53$. As far as the network design is concerned, we use a $3$ layer network consisting of fully-connected layers for modeling the two network streams. As opposed to the PAN-MS case where the input data dimensionality are different, both the streams, in this case, follows identical architecture.

\subsection{Experiments protocols}
For the case of PAN-MS data set, in Step-1, the PAN and MS network streams are individually trained with a learning rate or $3 \times 10^{-4}$. The same parameter values are also used to train the hallucination stream as well. For Step-2 where the initial two-stream network is trained and Step-4 where the hallucinated two-stream network is trained, the learning is performed with a learning rate of $1.5 \times 10^{-4}$. Each of the networks is trained for 100 iterations while the initial two-stream network is trained for 200 iterations. For experiments on HS data, each of the two streams and hallucination stream is learned with a learning rate of $6 \times 10^{-4}$ and the initial and hallucinated two-stream networks are learned using a learning rate of $3 \times 10^{-4}$. 

The performance is measured using the overall and class-wise accuracies of classification. We would like to highlight that we would ideally like the performance of Hallucination network to be as close to the performance of networks stream for the missing modality. Similarly, we would like the performance of the hallucinated two-stream network to be higher than that of the available modality stream but of course as close as possible to the original two-stream network trained with both the data modalities.

We also extensively study the effects of various parameters in modeling our network, namely, $\alpha$, $\lambda$ and $T$ along with the network depth.

\begin{table}[!t] 
\caption{\label{tab:expt1}Performance of different networks (in percentages) on PAN-MS dataset. The mean and standard deviation of overall accuracy over 10 runs is provided.}
\centering
\begin{tabular}{|p{1.8cm}|@{}c@{}|c|}
\hline
Classifier & Class-wise Accuracy & Overall Accuracy\\
\hline
PAN-Net & 
\begin{tabular}{c}
Aquafarm : 96.64\\\hline Cloud : 99.36\\\hline Forest : 94.13\\\hline High building : 95.48\\\hline Low building : 94.86 \\\hline Farm land : 90.36\\\hline River : 94.29\\\hline Water : 99.46 
\end{tabular}
&  95.82 (0.20)\\
\hline
MS-Net & 
\begin{tabular}{c}
Aquafarm : 95.27\\\hline Cloud : 99.87\\\hline Forest : 95.52\\\hline High building : 95.04\\\hline Low building : 98.00\\\hline Farm land : 92.41\\\hline River : 94.59\\\hline Water : 99.46 
\end{tabular}
& 96.16 (0.31)\\
\hline
Two-Stream \newline Net & 
\begin{tabular}{c}
Aquafarm : 96.03\\\hline Cloud : 99.59\\\hline Forest : 96.84\\\hline High building : 96.82\\\hline Low building : 98.57\\\hline Farm land : 95.13\\\hline River : 93.51\\\hline Water : 99.88 
\end{tabular}
& 98.23 (0.15)\\
\hline
\end{tabular}
\end{table}

\begin{table}[!t]
\caption{\label{tab:expt2}Performance of different networks (in percentages) on PAN-MS dataset while hallucinating each of the streams. The mean and standard deviation of overall accuracy over 10 runs is provided.}
\centering
\begin{tabular}{|p{1.8cm}|@{}c@{}|c|}
\hline
Classifier & Class-wise Accuracy & Overall Accuracy\\
\hline
MS \newline Hall-Net \newline (T=10) & 
\begin{tabular}{c}
Aquafarm : 97.27\\\hline Cloud : 99.12\\\hline Forest : 97.24\\\hline High building : 99.24\\\hline Low building : 92.38\\\hline Farm land : 90.59\\\hline River : 96.21\\\hline Water : 99.90 
\end{tabular}
& 96.12 (0.03)\\
\hline
MS \newline Hallucinated \newline Two-Stream Net & 
\begin{tabular}{c}
Aquafarm : 96.30\\\hline Cloud : 99.46\\\hline Forest : 96.32\\\hline High building : 96.65\\\hline Low building : 94.68\\\hline Farm land : 93.47\\\hline River : 97.30\\\hline Water : 100 
\end{tabular}
& 96.78 (0.08)\\
\hline
PAN \newline Hall-Net \newline (T=10) & 
\begin{tabular}{c}
Aquafarm : 95.53\\\hline Cloud : 99.67\\\hline Forest : 97.20\\\hline High building : 94.83\\\hline Low building : 97.90\\\hline Farm land : 92.79\\\hline River : 93.71\\\hline Water : 99.72 
\end{tabular}
& 96.42 (0.22)\\
\hline
PAN \newline Hallucinated \newline Two-Stream Net & 
\begin{tabular}{c}
Aquafarm : 96.65\\\hline Cloud : 99.91\\\hline Forest : 97.36\\\hline High building : 95.21\\\hline Low building : 98.66\\\hline Farm land : 94.18\\\hline River : 94.10\\\hline Water : 99.90 
\end{tabular}
& 97.01 (0.11)\\
\hline
\end{tabular}
\end{table}
\subsection{Results on PAN-MS Data}
We perform our experiments on the panchromatic and multi-spectral pair dataset with the convolutional network architecture as shown in figure \ref{fig:fullArch}. We use different training ratios for our experiments and study the performance variations with respect to them. The results are shown in Table \ref{tab:expt1} and Table \ref{tab:expt2} for hallucinating multi-spectral and panchromatic streams, while considering 50\% of the entire data for training, from which 5\% is used for validation. In these tables, the PAN-Net and MS-Net refer to the individual network streams for each of the modalities as in Step-1. The Two-stream Net is the network from Step-2, where the two individual streams are fused. The MS Hall-Net and PAN Hall-Net refer to the Hallucination networks from Step-3 for hallucinating Panchromatic and multi-spectral data streams respectively. The proposed Hallucinated two-stream Net is the two-stream network from Step-4 comprising of the hallucination stream in place of the missing modality. The table \ref{tab:expt2} shows overall and class-wise accuracies of each of the streams with hallucination, from where it is evident that the multi-spectral hallucination network has indeed learned to approximate the missing modality and achieves a comparable accuracy as its teacher network which is the original multi-spectral stream. The results presented are mean and standard deviation of the of the values.

And also the hallucinated two-stream network constructed with this hallucination stream gives us a higher overall accuracy and class-wise accuracies in comparison to the panchromatic stream. Similar observations can be made from table \ref{tab:expt2} about the panchromatic hallucination network which gives higher overall accuracy than the multi-spectral stream alone. Note that the performance of the two-stream network trained in Step-2 acts as the upper bound to what the hallucinated two-stream network is expected to achieve.

We can observe from results in table \ref{tab:expt2} that for multi-spectral data hallucination, out of the eight classes, the Cloud class is best classified by the each of the streams with an accuracy going over $99\%$ in each case. The Water class, on the other hand, is best hallucinated by the Hallucination Net and the hallucinated two-stream network achieves a performance close to the original two-stream accuracy for it. Also, we can observe that the class farmland is the most difficult to classify and is extremely difficult to hallucinate as well with its classification accuracy being the least among the available classes but still has a performance jump of around $+1\%$ in the hallucinated two-stream network in comparison to the PAN-Net. For panchromatic data hallucination from results in table \ref{tab:expt2}, the Cloud class is still best classified by all the networks, while the class River remains most difficult to hallucinate having a lower classification accuracy than even the Farm land class. The classification of the Farmland class suffers in this case too as the performance of the initial PAN and MS network streams is far from being good for it in particular. For this case, the jump in classification accuracy of Farmland class is around $4\%$ from MS-Net to hallucinated two-stream model.

\begin{figure}[!t]
\centering
\begin{subfigure}[b]{0.44\textwidth}
   \includegraphics[width=1\linewidth]{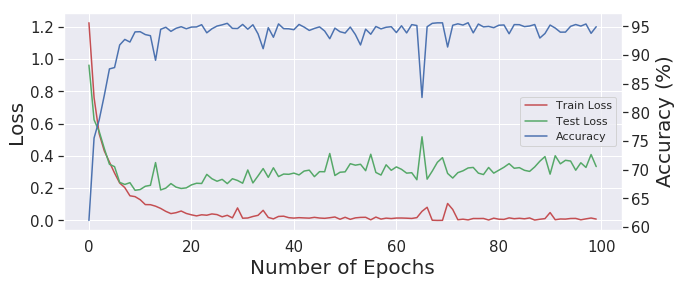}
   \caption{}
   \label{fig:metics1} 
\end{subfigure}
\begin{subfigure}[b]{0.44\textwidth}
   \includegraphics[width=1\linewidth]{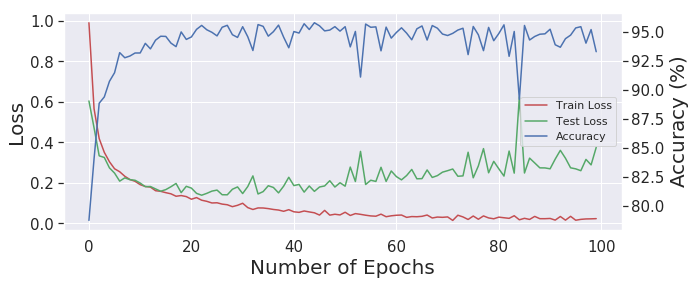}
   \caption{}
   \label{fig:metrics2}
\end{subfigure}
\begin{subfigure}[b]{0.44\textwidth}
   \includegraphics[width=1\linewidth]{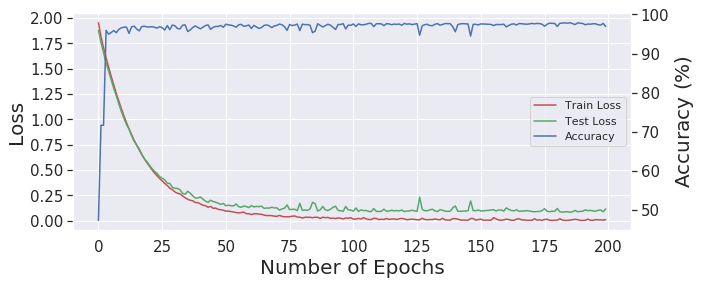}
   \caption{}
   \label{fig:metics3} 
\end{subfigure}
\begin{subfigure}[b]{0.44\textwidth}
   \includegraphics[width=1\linewidth]{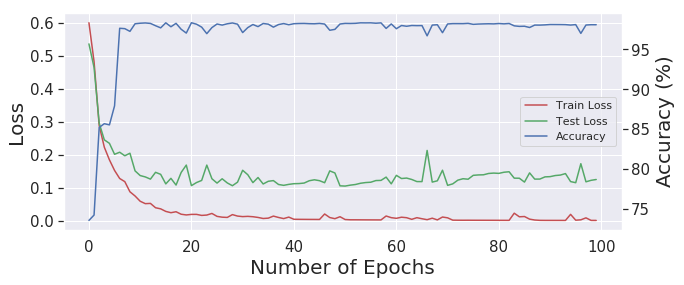}
   \caption{}
   \label{fig:metrics4}
\end{subfigure}
\caption{Training performance curves for (a) Panchromatic network, (b) Multi-spectral network, (c) Two-Stream network and (d) Final Hallucination network.}
\label{fig:trainingCurves}
\end{figure}

We further present the error and accuracy curves of the four networks while training namely, PAN-net, MS-net, Two-Stream Net and the hallucinated two-stream net, for 100, 100, 200 and 100 training epochs, respectively. We consider the best performing network snapshots over validation set during each of these training processes for deployment and use in subsequent stages. It can in Figure \ref{fig:trainingCurves} be observed that the final network converges quickly since it merely requires slight fine-tuning which is performed at a lower learning rate in comparison to the other networks.

\subsection{Abalation Study}
\subsubsection{Effect of varying network parameters}
\begin{figure}[!t]
\centering
\includegraphics[scale=0.39]{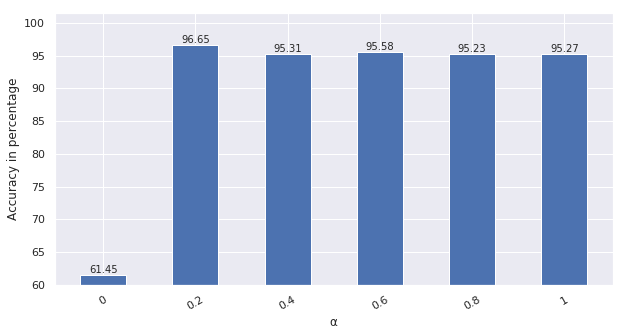}
\caption{Effect of $\alpha$ on classification accuracy (in percentages) of Hallucination network on PAN-MS dataset.}
\label{fig:alphaAbalation}
\end{figure}

We subsequently conduct experiments to study the effect of changing various hyper-parameters of our model and also consider different training-test splits in order to assess the learning capability of our network under different levels of supervision. We defined $\alpha$ to be the weighing factor of knowledge distillation loss in the overall loss term. Figure \ref{fig:alphaAbalation} shows how by increasing the parameter $\alpha$, the accuracy of the hallucination network increases first and then decreases for higher values of $\alpha$. From this, we can easily infer the advantage of the inclusion of the knowledge distillation loss term in our objective function. In particular, $\alpha = 0$ implies only the hallucination loss term is used and $\alpha =1$ is for when only the knowledge distillation loss term is prioritized. We can observe from figure \ref{fig:alphaAbalation} that the knowledge distillation loss results in an improved hallucination network. As desired, a weighted combination of these two loss measures provides an improved performance in terms of classification accuracy in comparison to just using one of them. Essentially, we observe a substantial improvement in the learning capability of the hallucination network by using the proposed generalized distillation technique instead of resorting to only the hallucination loss.

\begin{figure}[!t]
\centering
\includegraphics[scale=0.39]{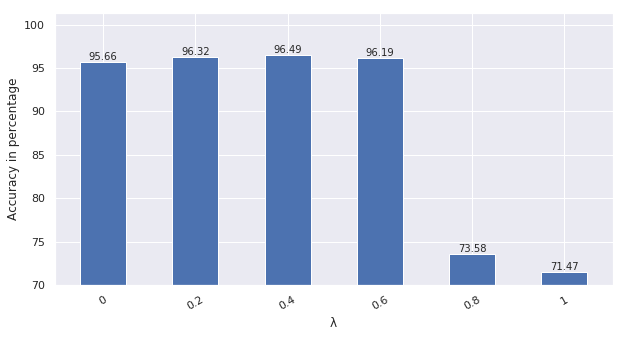}
\caption{Effect of $\lambda$ on classification accuracy (in percentages) of Hallucination network on PAN-MS dataset.}
\label{fig:lambdaAbalation}
\end{figure}

In figure \ref{fig:lambdaAbalation} we can observe how the increase in the value of $\lambda$ parameter changes the learning of our hallucination stream. As a part of the knowledge distillation loss, the parameter $\lambda$ weighs the constituent losses of the $L_{KD}$. If $\lambda = 0$, only soft threshold are used for knowledge distillation and in the case of $\lambda = 1$, only the hard targets are used for the distillation process. We can observe that soft targets give us a better hallucination performance in comparison to only the hard targets. In this case, too, we obtain a much better hallucination network by using a weighted combination of these losses instead of just resorting to using hard targets for distillation loss.

\subsubsection{Effect of varying temperature}
\begin{figure}[!t]
\centering
\includegraphics[scale=0.39]{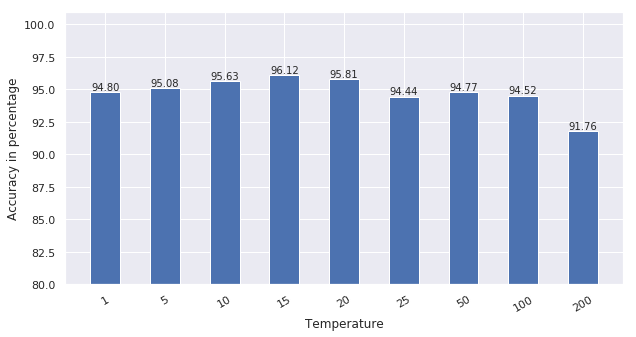}
\caption{Effect of $T$ on classification accuracy (in percentages) of Hallucination network on PAN-MS dataset.}
\label{fig:temperatureAbalation}
\end{figure}

Although the temperature parameter is difficult to tune we found a trend in the same on averaging the results across multiple runs. We repeated the proposed MS hallucination process for multiple temperature values ranging from 1 to 100 while $\alpha$ and $\lambda$ were each set to $0.5$. The results of this are presented in Fig. \ref{fig:temperatureAbalation}. We can observe from the figure that the soft threshold obtained at $T=15$ give best hallucination performance for this particular case. Another observation is that the performance initially increases with increase in temperature but quickly starts dropping with further increase.

\begin{table}[!t]
\caption{\label{tab:expt4}Performance of different networks on PAN-MS dataset (in percentages) on reducing network depth by one each of convolution and fully connected layer in each of the network streams.}
\centering
\begin{tabular}{|p{5cm}|c|}
\hline
Classifier & Overall accuracy\\
\hline
PAN-Net & 93.97\\
\hline
MS-Net & 95.33\\
\hline
Two-Stream Net & 96.43\\
\hline
MS Hall-Net (T=10) & 93.74\\
\hline
Hallucinated Two-Stream Net & 95.50\\
\hline
\end{tabular}
\end{table}

\subsubsection{Effect of varying stream depth}
The next experiment involves varying the capacity of each our streams. Specifically,  we reduce the number of layers in each network stream and observe how this affects the hallucination and hallucinated two-stream network performance. The corresponding performance measures are tabulated in table \ref{tab:expt4}. We observe that reducing the network depth affects the learning capacity of the hallucination network along with all of the other networks. The results are shown for the case of hallucinating multi-spectral data stream. We observe that the performance of initial MS net drop by 1.5\% in comparison to the results from table \ref{tab:expt1} and that of the hallucination net drops by 3\% due to the reduction in network parameters.

\subsubsection{Effect of varying training data size}
For this study, we use three different training-test splits to carry out the training process and report the observed changes in final classification performance. In particular, we use training sets comprising of 20\%, 50\% and 80\% samples from the entire dataset. Table \ref{tab:valAbalation} shows the accuracies for varying training ratios for each of our networks, where we hallucinate the multi-spectral data stream. As expected, it can be observed that the accuracies of each of the networks increase with an increase in training set size. The hallucination network also improves with an increase in data and as a result, the hallucinated two-stream network always showcase a higher accuracy measure than considering solely the PAN-stream.

\begin{table}[!t]
    \centering
        \caption{\label{tab:valAbalation}Performance of different networks on PAN-MS dataset (in percentages) for varying training sizes.}
    \begin{tabular}{c|c|c|c|}
        \cline{2-4}
         & \multicolumn{3}{|c|}{Training Ratio} \\
        \cline{2-4}
         & 20\% & 50\% & 80\%\\
        \hline
        \multicolumn{1}{|c|}{PAN-Net} & 95.05 & 95.60 & 96.50\\
        \hline
        \multicolumn{1}{|c|}{MS-Net} & 93.95 & 96.43 & 96.60\\
        \hline
        \multicolumn{1}{|c|}{Two-Stream Net} & 95.51 & 97.97 & 98.32\\
        \hline
        \multicolumn{1}{|c|}{MS Hall-Net (T=10)} & 92.78 & 96.20 & 96.48\\
        \hline
        \multicolumn{1}{|c|}{Hallucinated Two-stream Net} & 94.63 & 96.87 & 97.79\\
        \hline
    \end{tabular}
    \label{table::supervised1}
\end{table}

\subsection{Results on Hyperspectral Data}
We divide the Pavia University hyperspectral datacube into two parts and used them to train each of the streams in order to hallucinate the other half of the bands in the datacube. The same train-test split is employed here as the PAN-MS data. The results for hallucinating the second (53) and the first half (52) of the bands are shown in table \ref{tab:expt3}, respectively.

\begin{table}[!t]
\caption{\label{tab:expt3}Performance of different classifiers on Pavia hyperspectral datacube (in percentages) while hallucinating the first 52 and last 53 bands. The mean and standard deviation of overall accuracy over 10 runs is provided.}
\centering
\begin{tabular}{|p{5cm}|c|}
\hline
Classifier & Overall accuracy\\
\hline
Stream one & 85.29 (0.14)\\
\hline
Stream two & 93.57 (0.22)\\
\hline
Two-stream Net & 95.82 (0.15)\\
\hline
Hallucination network for \newline band 53-103, (T=15) & 85.12 (0.17)\\
\hline
Hallucinated Two-stream Net \newline for band 53-103 & 86.04 (0.25)\\
\hline
Hallucination network for \newline band 1-52, (T=15) & 93.33 (0.07)\\
\hline
Hallucinated Two-stream Net \newline for band 1-52 & 94.24 (0.12)\\
\hline
\end{tabular}
\end{table}

From the tables, it can be noted that these two sub-cubes offer fairly complementary contributions to the final classification accuracy. Similar to the case of PAN-MS dataset, we can see that the hallucination network has successfully learned to mimic the missing bands of the datacube to a considerable extent. Besides. we can also observe that the hallucinated two-stream network offers an improved performance in comparison to the single stream networks trained on either half the datacube. This indicates that we can obtain an improved performance with a lesser number of bands using our hallucination architecture than a unimodal classifier.

\section{Conclusions}
We propose a hallucination network model in this paper which can deal with the problem of missing data modality during testing in a typical remote sensing classification task. It is imperative that information fusion from multiple sources provide complementary insights to a geographical area which helps in better characterization. However, in many situations, some of the modalities may remain absent. Our approach is capable of approximating such missing modalities during the inference stage. Ideally, we train our model with all the available modalities, thus giving rise to the notion of learning with privileged information. We validate the proposed framework on a large-scale PAN-MS dataset for scene recognition and a hyperspectral image for pixel classification where the effectiveness of model hallucination is established through extensive experimentation. We plan to extend this model to handle multi-temporal data as a future endeavor.


\bibliographystyle{IEEEbib}
\bibliography{refs}

\end{document}